 \documentclass[final,3p,times]{elsarticle}

\usepackage{amssymb}
 \usepackage{multirow} 
\usepackage{dirtytalk}
\usepackage{graphicx}
\usepackage{array}
\usepackage{url}
\usepackage{subcaption}
\usepackage{amsmath}

\journal{Knowledge-Based Systems}

\begin{document}

\begin{frontmatter}



\title{Evaluating Sensitivity Parameters in Smartphone-Based Gaze Estimation: A Comparative Study of Appearance-Based and Infrared Eye Trackers}


\author[a,c]{Nishan Gunawardena\corref{cor1}} 
\ead{n.gunawardena@unsw.edu.au}
\cortext[cor1]{Corresponding author}

\author[b]{Gough Yumu Lui}
\ead{G.Lui@westernsydney.edu.au}

\author[a]{Bahman Javadi}
\ead{b.javadi@westernsydney.edu.au}

\author[a]{Jeewani Anupama Ginige}
\ead{j.ginige@westernsydney.edu.au}

\address[a]{School of Computer, Data and Mathematical Sciences, Western Sydney University, Penrith, New South Wales, Australia}
\address[b]{The MARCS Institute for Brain, Behaviour and Development, Western Sydney University, Penrith, NSW, Australia}
\address[c]{School of Systems and Computing, University of New South Wales, Canberra, ACT, Australia}

\begin{abstract}

This study evaluates a smartphone-based, deep-learning eye-tracking algorithm by comparing its performance against a commercial infrared-based eye tracker, the Tobii Pro Nano. The aim is to investigate the feasibility of appearance-based gaze estimation under realistic mobile usage conditions. Key sensitivity factors, including age, gender, vision correction, lighting conditions, device type, and head position, were systematically analysed. The appearance-based algorithm integrates a lightweight convolutional neural network (MobileNet-V3) with a recurrent structure (Long Short-Term Memory) to predict gaze coordinates from grayscale facial images. Gaze data were collected from 51 participants using dynamic visual stimuli, and accuracy was measured using Euclidean distance. The deep learning model produced a mean error of 17.76 mm, compared to 16.53 mm for the Tobii Pro Nano. While overall accuracy differences were small, the deep learning-based method was more sensitive to factors such as lighting, vision correction, and age, with higher failure rates observed under low-light conditions among participants using glasses and in older age groups. Device-specific and positional factors also influenced tracking performance. These results highlight the potential of appearance-based approaches for mobile eye tracking and offer a reference framework for evaluating gaze estimation systems across varied usage conditions.

\end{abstract}

\begin{keyword}



Eye tracking \sep Smartphones \sep Sensitivity Analysis  \sep Performance Benchmarking  \sep Deep Learning for Eye Tracking  \sep Gaze Estimation  \sep Human-Computer Interaction.

\end{keyword}

\end{frontmatter}




\section{Introduction}
\label{intro}

The growing use of smartphones with front-facing cameras has opened new possibilities for gaze-based human-computer interaction. Eye-tracking technology, traditionally limited to specialised hardware such as infrared-based commercial eye trackers, is now being adapted to work on mobile devices using deep learning-based appearance models. This shift enables a range of applications, such as accessibility tools, gaming, user attention analysis, and mobile health diagnostics, to operate without requiring dedicated equipment \cite{gunawardena2022}.

The global eye-tracking market has been experiencing rapid growth. For instance, the market size was valued at USD 850 million in 2023 and is anticipated to grow at a compound annual growth rate (CAGR) of over 30.5\% between 2024 and 2032 \cite{gminsights2023eye}. This growth is driven by increasing adoption in sectors such as healthcare, gaming, and marketing research, where understanding user behaviour is crucial. Despite the growing interest, smartphone-based eye tracking remains a technically challenging task due to varying environmental conditions, hardware limitations, and user diversity. Specific challenges include limited camera resolution, inconsistent device orientation, varying lighting conditions, and demographic factors such as age, vision correction, and facial structure. Each of these can significantly affect the algorithm’s accuracy, making comprehensive evaluation in real-world scenarios essential.

Over the past decade, researchers have assessed the performance of various appearance-based eye-tracking algorithms against model-based methods and commercial eye-trackers in real-world applications. Zhang et al. \cite{zhang2019evaluation} evaluated GazeML \cite{park2018learning} (model-based), MPIIFaceGaze \cite{zhang2017s}(appearance-based method) and Tobii EyeX \cite{tobii_eyex} (commercial eye-tracker) to compare the accuracy across various conditions, including different distance between user and camera, calibration methods, indoor/outdoor settings and the use of prescription glasses. The results of this study suggest that the appearance-based method achieved higher accuracy across different distances and prescription glasses, outperforming the model-based method. This finding highlights the practicality of using appearance-based gaze estimation in various conditions. To assess the feasibility of deploying gaze estimation in mobile environments, it is important to compare appearance-based methods with infrared-based commercial solutions, which are typically considered more accurate but less accessible. Such comparative analysis allows researchers to identify the trade-offs between performance, cost, and scalability.

Although prior research has examined appearance-based methods in mobile contexts, many studies focus on isolated variables such as calibration or distance from the device. Comprehensive evaluations that consider a broader range of sensitivity parameters such as age, gender, use of vision correction, device type, and head position are limited. Moreover, there is a lack of direct comparisons between mobile appearance-based approaches and commercial infrared-based systems using real-world data. This study aims to address that gap by evaluating a deep learning-based gaze estimation algorithm developed for mobile devices~\cite{gunawardena2024deep}, using the Tobii Pro Nano eye tracker~\cite{tobii_pro_nano} as a reference. The evaluation focuses on how different user and context-specific variables influence gaze estimation accuracy, providing insights into the conditions under which mobile-based gaze estimation performs reliably. The Tobii Pro Nano serves as a baseline due to its established role in eye-tracking research and its use of infrared technology for gaze detection.

This paper makes the following contributions:

\begin{itemize}
  \item Evaluates the impact of user and context-specific factors, including age, gender, vision correction, lighting conditions, device type, and head position, on the performance of appearance-based gaze estimation systems.

  \item Presents a comparative case study between a deep learning-based mobile eye-tracking method and a commercial infrared-based tracker, identifying performance characteristics and trade-offs.

  \item Provides practical recommendations for the design of generalisable, calibration-free eye-tracking systems intended for deployment in real-world mobile environments.
\end{itemize}

The remainder of this paper is structured as follows. Section~\ref{related_works} reviews related methodologies and existing approaches in mobile device eye-tracking research. Section~\ref{mobileeye} presents the proposed deep learning-based architecture. Section~\ref{sec:method} details the experimental setup, implementation procedures, and data collection process. Section~\ref{sec:results} discusses the evaluation results and key findings. Finally, Section~\ref{conclusion} concludes the paper and outlines directions for future work.

\section{Related Works}
\label{related_works}

Eye tracking has traditionally relied on infrared-based, model-driven techniques that require dedicated hardware and careful calibration \cite{tobii_eyex}. These systems offer high precision and are widely used in controlled environments, but their cost, portability limitations, and dependency on specific calibration procedures make them less suitable for large-scale or mobile applications. In contrast, appearance-based gaze estimation methods leverage RGB images captured from regular cameras and use machine learning models, particularly convolutional neural networks—to infer gaze direction \cite{zhang2017s}. This paradigm has gained traction due to its ability to operate on consumer-grade hardware, such as laptops and smartphones, without the need for infrared illumination or specialised sensors.

Recent advances in appearance-based gaze estimation have enabled the development of models that work on mobile devices, using only the front-facing camera. The objective of appearance-based gaze estimation is to learn a direct mapping from facial images to gaze coordinates. Driven by improvements in front-facing camera quality and the availability of large-scale gaze datasets, several studies have demonstrated the feasibility of deploying such models on mobile devices. Most approaches are based on convolutional neural networks (CNNs), which are well suited for extracting spatial features from visual input \cite{lei2023end}. For instance, Krafka et al. \cite{krafka2016eye} introduced a CNN-based model using face and eye crops to estimate gaze on mobile devices, while B{\^a}ce et al. \cite{bace2019accurate} proposed a two-stage CNN pipeline for face detection and gaze estimation. Valliappan et al. \cite{valliappan2020accelerating} refined the CNN-based approach by incorporating a calibration step and additional regression models to improve accuracy. Moreover, previous work has evaluated different lightweight CNN architectures, such as MobileNet \cite{howard2017mobilenets} and Shufflenet \cite{zhang2018shufflenet}, for smartphone-based eye tracking \cite{gunawardena2022performance}. Although most of these models were trained and evaluated on the GazeCapture dataset \cite{krafka2016eye}, they largely focus on model development under limited evaluation settings.

Further, these methods were primarily evaluated using static visual stimuli—such as fixed dots or still images where the gaze target remains stationary. In contrast, dynamic stimuli involve continuous movement of visual elements, such as moving dots or video content, which better reflect real-world use cases like gaming, video-based learning, and mobile app interactions. To address this, our previous study \cite{gunawardena2024deep} proposed a smartphone-based eye-tracking algorithm that integrates CNNs with recurrent neural networks (RNNs) to capture both spatial and temporal information from video sequences. This hybrid architecture was specifically designed to estimate gaze during dynamic visual interactions, providing a more realistic benchmark for mobile device eye tracking. Section \ref{mobileeye} will further discuss this architecture and its accuracy.

While several studies have compared smartphone-based eye-tracking models with commercial solutions, most have focused on specific evaluation aspects. For instance, Strobl et al.~\cite{strobl2019look} examined calibration strategies in the iTracker model, and Valliappan et al.~\cite{valliappan2020accelerating} evaluated their CNN-based approach against Tobii Pro Glasses 2 under controlled conditions. Other efforts, such as those by Zhang and Cui~\cite{zhang2022reliability}, explored the application of Tobii Pro Nano in cognitive workload tasks, while Le et al.~\cite{le2022practical} integrated commercial eye tracking with a mobile app to assess performance across different user postures. These studies contribute to understanding model performance under various constraints but are often tailored to structured tasks or static viewing scenarios.

Building on this prior work, there is value in expanding the scope of evaluations to include a wider range of user and environmental factors. Elements such as lighting, vision correction, head position, and age may influence the performance of appearance-based gaze estimation models~\cite{swift2021facial, agbolade2018two, paulus2017use, wyder2018eye, marquardt2024selection}. A systematic investigation that includes these factors can support further development of eye-tracking systems that operate under everyday conditions. This study contributes to this area by examining a set of such parameters in the context of mobile-based gaze estimation and comparing the outcomes with a commercial infrared-based solution.

This study addresses that need by evaluating the performance of a deep learning-based gaze estimation algorithm, previously developed for dynamic visual stimuli~\cite{gunawardena2024deep}, in comparison with the Tobii Pro Nano~\cite{tobii_pro_nano}. The evaluation is conducted across a wide range of sensitivity parameters using real user data, providing a broader understanding of the algorithm’s applicability and robustness in practical deployment scenarios.

\section{MobileEYE Architecture Overview}
\label{mobileeye}

This study uses a deep learning-based gaze estimation model designed for real-time mobile deployment. The architecture integrates MobileNet-V3 as the convolutional backbone for spatial feature extraction, followed by Long Short-Term Memory (LSTM) layers to model temporal dependencies. The model predicts gaze coordinates from $128 \times 128$ grayscale face images captured using a smartphone’s front-facing camera. The choice of MobileNet-V3 was informed by prior comparative evaluations on lightweight CNNs for mobile gaze estimation~\cite{gunawardena2022performance}, where it offered an optimal trade-off between accuracy and efficiency. LSTM layers were incorporated to handle the temporal nature of dynamic visual stimuli such as video clips, as detailed in earlier work~\cite{gunawardena2024deep}. This architecture, originally proposed in our previous work~\cite{gunawardena2024deep}, will be referred to as \textbf{MobileEYE} throughout this paper.

\begin{figure}[htbp]
    \centering
    \includegraphics[width=0.8\textwidth]{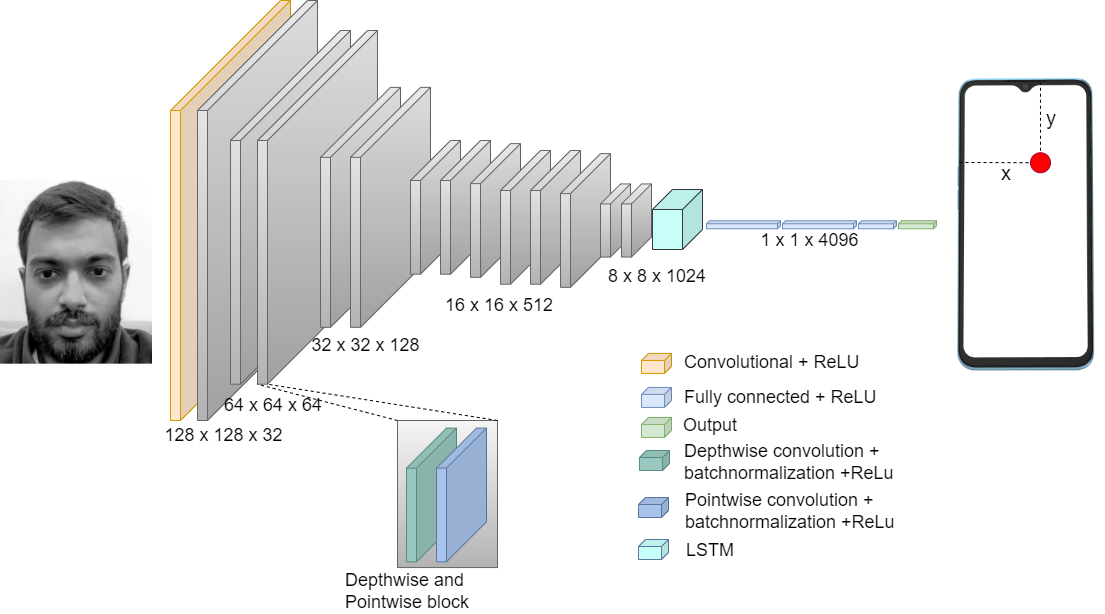}
    \caption{CNN+LSTM architecture for smartphone-based gaze estimation.}
    \label{fig:mobileeye_architecture}
\end{figure}

The model was optimised using TensorFlow Lite for deployment on mobile and edge devices. On-device inference was tested on a Samsung S22, achieving an average latency of 426 ms per frame~\cite{gunawardena2025smartphone}. Figure~\ref{fig:mobileeye_architecture} illustrates the end-to-end CNN+LSTM structure used in this study. Unlike prior models such as iTracker~\cite{krafka2016eye}, which rely on multiple input streams (e.g., eyes, face grid), this architecture uses only the face image, reducing computational complexity and enabling streamlined inference suitable for real-time applications on smartphones.

\section{Evaluation Methodology}
\label{sec:method}

This section outlines the methodology adopted for evaluating the MobileEYE algorithm. It includes details of the experimental setup, device selection, implementation strategies, and data collection procedures.

\subsection{Experimental Setup}

The experimental setup was designed to evaluate the accuracy and robustness of the MobileEYE algorithm under real-world conditions, comparing its predictions against a commercial eye tracker. It involves the use of two smartphones, a screen-based eye tracker, a local server, and a web-based application for video stimulus presentation and data collection. The setup also includes various environmental configurations, such as lighting conditions and different device-holding positions, to assess the algorithm's performance in diverse scenarios.

\subsubsection{Device and Eye Tracker Selection}

Two smartphones, the Samsung Galaxy S22 and the iPhone 14, were used for data collection. These devices were selected due to their similar screen sizes, modern hardware capabilities, and relevance to current user trends. Both phones were released in 2022 and include front-facing cameras capable of high-resolution video recording, enabling consistent data capture. The use of two distinct mobile platforms also allowed for examining the effect of device-specific hardware on gaze estimation performance.

Several commercial eye trackers were considered for comparison. Initial tests involved Tobii Pro Glasses 3, a wearable tracker previously used in mobile gaze estimation studies~\cite{valliappan2020accelerating}. However, limitations were identified, including calibration challenges for smartphone interactions, restricted visibility of the screen due to field of view constraints, and incompatibility with evaluating users wearing prescription glasses. As a result, a screen-based eye tracker, Tobii Pro Nano, was used with a fixed mobile phone stand. This setup allowed consistent reference between the device screen and the eye tracker, improving gaze point alignment during data collection. Tobii Pro Nano was the only compatible commercial screen-based tracker available during the experimental phase and was used as the baseline system in this study.

\subsubsection{Participant Positions}
\label{three_positions}

\begin{figure}[ht]
    \centering
    \begin{subfigure}[b]{0.28\textwidth}
        \centering
        \includegraphics[width=\textwidth]{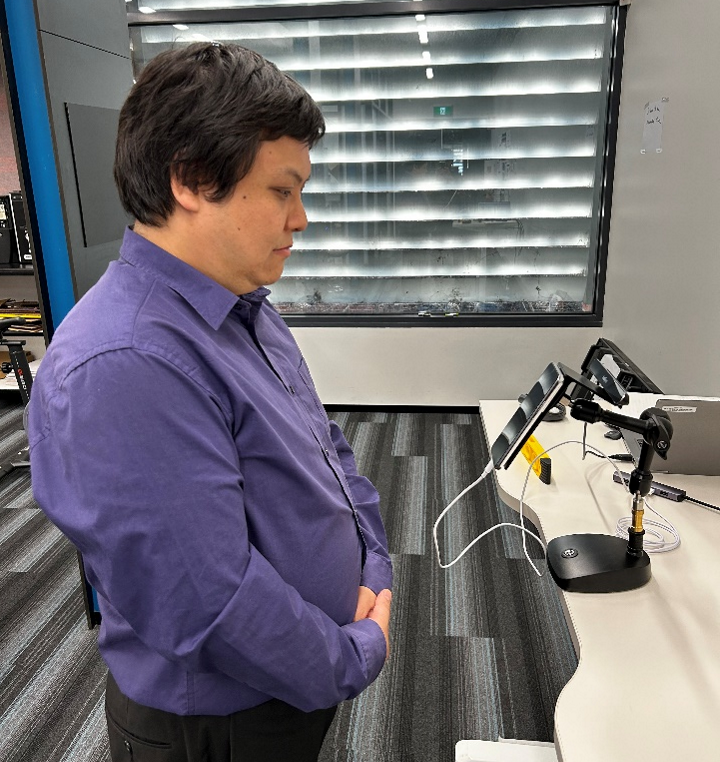}
        \caption{Position 1: Looking down.}
        \label{fig:image1}
    \end{subfigure}
    \hfill
    \begin{subfigure}[b]{0.28\textwidth}
        \centering
        \includegraphics[width=\textwidth]{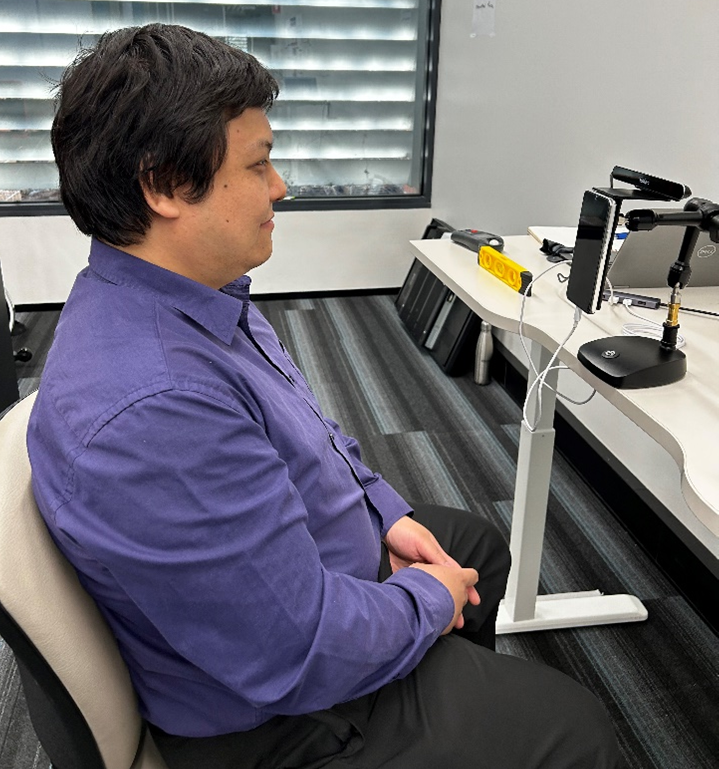}
        \caption{Position 2: Looking straight.}
        \label{fig:image2}
    \end{subfigure}
    \hfill
    \begin{subfigure}[b]{0.28\textwidth}
        \centering
        \includegraphics[width=\textwidth]{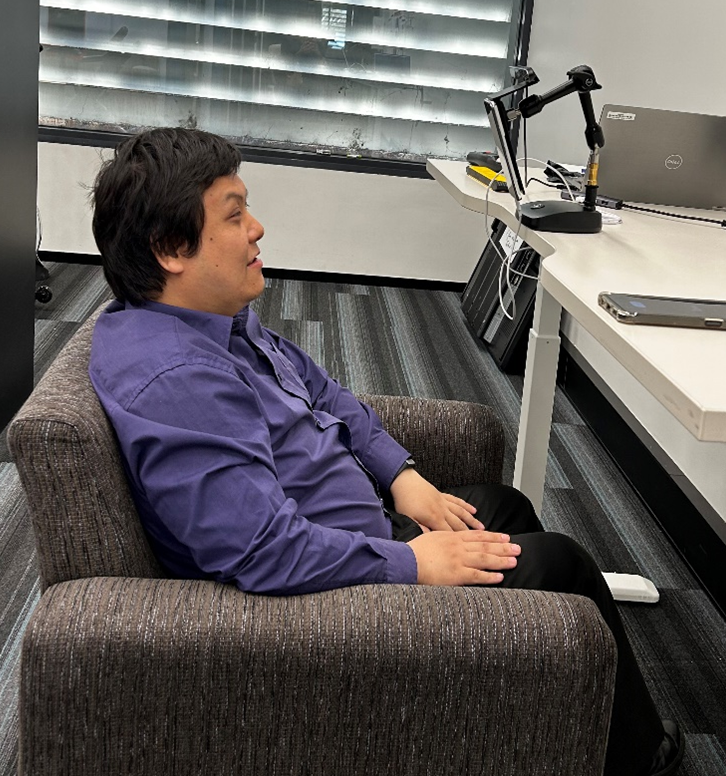}
        \caption{Position 3: Looking up.}
        \label{fig:image3}
    \end{subfigure}
    \caption{Three distinct head positions used in this experiment}
    \label{fig:three_positions}
\end{figure}

In this experiment, participants’ eye movements were recorded in three distinct head positions: looking down, straight, and up (Figure~\ref{fig:three_positions}). These positions were selected to reflect common ways users interact with smartphones in everyday settings. Looking down represents postures such as reading or browsing while seated or standing, where the upper eyelid may partially occlude the eye. Looking straight corresponds to holding the device at eye level, such as during video calls or when the phone is mounted. Looking up simulates usage scenarios like lying down, where the lower eyelid may cover parts of the eye.

Tracking performance was evaluated across these positions to examine the effect of head posture on gaze estimation accuracy. Each position introduces different angles and visibility conditions for the eye, influencing the image input to the model. This assessment helps determine the consistency of gaze prediction across varying usage postures, providing insight into the applicability of the algorithm under typical mobile interaction conditions.

\subsection{Physical Setup}

\begin{figure}[ht]
    \centering
    \includegraphics[width=0.65\textwidth]{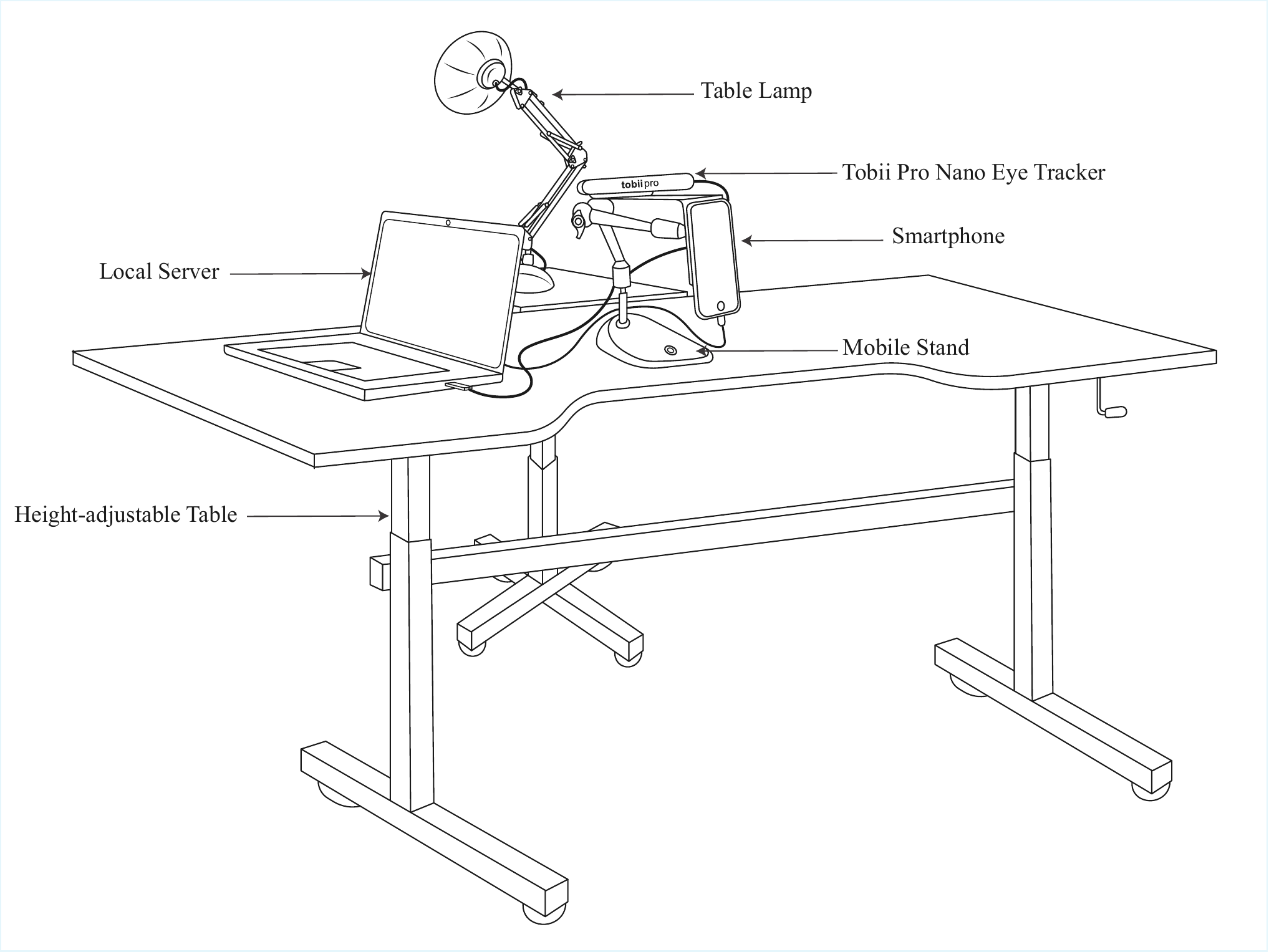}
    \caption{Physical Setup for Experiments}
    \label{fig:experimental_setup}
\end{figure}

Figure~\ref{fig:experimental_setup} shows the physical setup used for data collection, which included a local server (laptop), a Tobii Pro Nano eye tracker, a mobile phone stand, two smartphones, a table lamp, and a height-adjustable table. The Tobii Pro Nano was connected to the server via USB 3.0. The table height was adjusted for each participant to align the smartphone screen and eye tracker with the user’s eye level, supporting consistent data capture across participants.

Lighting conditions were controlled to simulate realistic use cases. In low-light settings, all overhead lights were turned off, and a single table lamp was used as the primary light source. The lamp was positioned 150 cm in front of the smartphone setup and directed toward a black screen to minimise reflections. An additional light source behind the participant was present, reducing the contrast between the eye region and surrounding skin, which affected image clarity. The setup was designed to assess algorithm performance under varied lighting conditions.

\begin{figure}[ht]
  \centering \includegraphics[width=\textwidth]{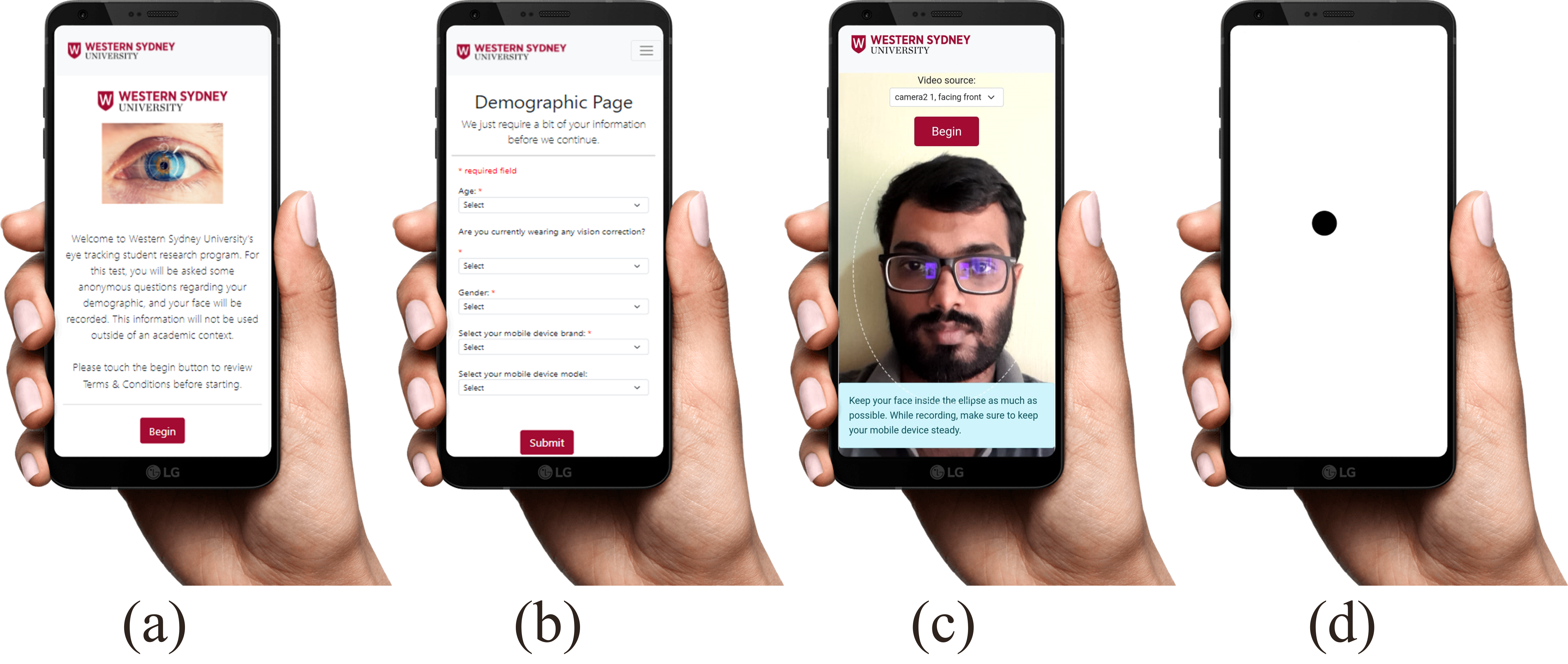}
  \caption{Web-based application developed to collect data through crowdsourcing. (a) Home screen, (b) Demographic information, (c) Camera selection, (d) Playing video}
  \label{fig:web-based application}
\end{figure}

Two applications were used for data collection. A web-based application was developed using Laravel PHP, JavaScript, and MySQL and deployed on both smartphones to present the visual stimuli and record interactions (Figure~\ref{fig:web-based application}). A second application, built with Flask and the Tobii Pro SDK in Python, ran on the local server to interface with the Tobii Pro Nano and record gaze data. For gaze calibration, a 5-point calibration method was used, where participants were instructed to look at a blinking dot shown at five locations on the smartphone screen—four corners and the centre. The screen size of each smartphone was defined in the application to align with the calibration routine required by the Tobii Pro Nano.

All devices were connected via a local network using an ASUS RT-AX86U Pro router without internet access, ensuring stable and fast communication between the server and smartphones during the experiment.

\subsubsection{Time Synchronisation Across Devices}

Synchronising time across multiple devices (two smartphones, a local server, and a commercial eye tracker) was essential to ensure alignment between gaze data and events on the smartphone screens. Each device operates on its internal clock, and variations can cause timestamp misalignments. Manual clock adjustment was not practical, and using an external Network Time Protocol (NTP) server was not feasible due to the offline local router setup. Alternative methods, such as synchronised start signals or post-processing alignment, were considered but presented challenges related to precision and complexity.

To address this, a latency compensation method was used. Before each recording session, 100 ping requests were sent from each smartphone to the local server to measure round-trip time (RTT). The average RTT was then used to adjust the smartphone timestamps by subtracting half of the RTT and aligning them with the server's time reference. The Tobii Pro SDK provided system timestamps from the eye tracker host machine, which were matched with the adjusted smartphone timestamps. This approach enabled consistent alignment of data recorded across all devices during the experiment.

\subsection{Data Collection Procedures}

Before collecting data, human ethics approval from Western Sydney University (Approval Number - H14493) was obtained to collect human-identifiable information, such as facial data. Human participants who use smartphones in their daily lives and are between 18 and 75 years old were considered for this experiment. An a priori power analysis was conducted to determine the number of participants needed for the study. The analysis considered a one-sample t-test with the following parameters: an effect size (d) of 0.39, an alpha error probability ($\alpha$) of 0.05 and a desired power (1-$\beta$) of 0.8. Based on these parameters, the analysis indicated that a total sample size of 43 participants is required to achieve the desired statistical power. A total of 51 participants were recruited for the evaluation. Each participant was expected to contribute six recordings per device, totalling 612 recordings. However, recordings from the Android devices of three participants (18 recordings) were not uploaded, and one participant was removed due to distraction during both sessions (12 recordings). As a result, 30 recordings were excluded, and the final dataset consisted of 582 recordings: 300 from the iPhone 14 and 282 from the Samsung S22. Table~\ref{table:participant_demographics} summarises the demographic characteristics of the 50 retained participants, including age group, gender, ethnicity, and vision correction status.

\begin{table}[ht]
\centering
\caption{Demographic summary of the 50 participants.}
\begin{tabular}{ll}
\hline
\textbf{Category} & \textbf{Count} \\
\hline
\multicolumn{2}{l}{\textit{Age Group}} \\
18--28 & 9 \\
29--38 & 22 \\
39--48 & 9 \\
49--58 & 6 \\
59--68 & 4 \\
\hline
\multicolumn{2}{l}{\textit{Gender}} \\
Male   & 25 \\
Female & 25 \\
\hline
\multicolumn{2}{l}{\textit{Ethnicity}} \\
South Asian        & 27 \\
Middle Eastern     & 6 \\
European           & 9 \\
Southeast Asian    & 4 \\
East Asian         & 3 \\
Hispanic/Latino    & 1 \\
\hline
\multicolumn{2}{l}{\textit{Vision Correction}} \\
None               & 30 \\
Glasses            & 18 \\
Contact Lenses     & 2 \\
\hline
\end{tabular}
\label{table:participant_demographics}
\end{table}

The order of head positions, lighting conditions, and smartphones was randomised for each participant to reduce potential order effects. Participants first submitted demographic details, including age, gender, and vision correction status, through the smartphone web application. They also completed a set of tutorial pages explaining the procedure. Before each recording session, participants selected one of the three head positions and a lighting condition (high or low light). They then performed calibration for the commercial eye tracker, and the ambient light level was measured using a digital lux meter placed parallel to the participant’s face to assess lighting conditions before recording began.

\begin{figure}[ht]
  \centering 
  \includegraphics[width=0.70\textwidth]{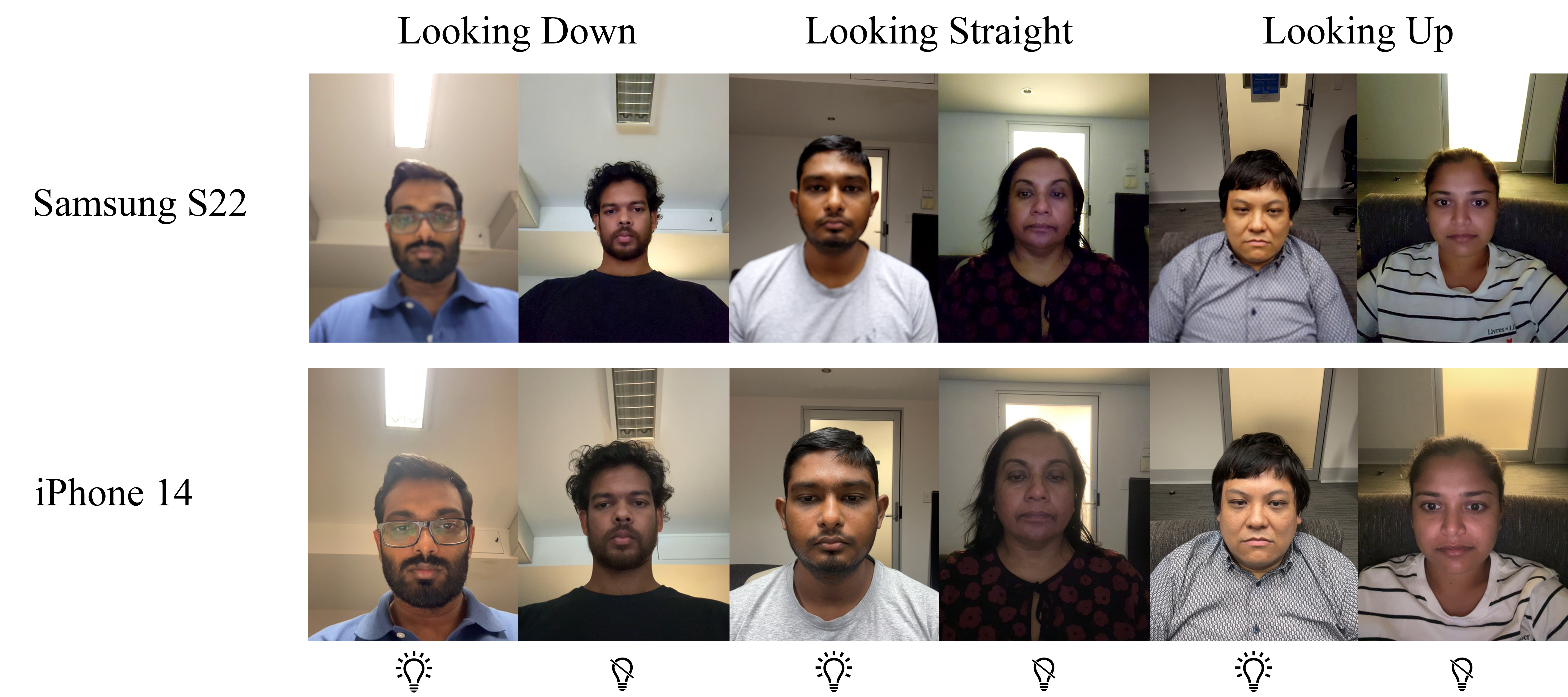}
  \caption{Sample frames from the collected data with two smartphones visualising three positions and two lighting conditions.}
  \label{fig:collected_data}
\end{figure}

Once calibration and light measurement were completed, participants followed a moving dot on the smartphone screen in full-screen mode. The dot's path and colour varied to introduce diverse gaze patterns. Each recording session was followed by an upload step, where the recorded video was sent to the server before participants could proceed. On average, the entire recording process lasted 21 minutes and 4 seconds, with a standard deviation of 4 minutes and 37 seconds. Figure~\ref{fig:collected_data} shows example frames from the collected dataset.

\section{Results and Discussion}
\label{sec:results}

Data collected in the above-mentioned lab setting included videos of the participants captured through the front-facing camera, coordinates of the stimuli, demographic data, calibration and gaze data from the Tobii Pro Nano eye tracker.

\subsection{Overall Error Analysis}

As the first step of the evaluation, the overall error of the Tobii Pro Nano eye tracker and the MobileEYE were evaluated to assess their performance and reliability. Table \ref{tab:overall_error} collates the results of the overall error analysis. The mean Euclidean distance for Tobii Pro Nano is 16.53 $\pm$ 7.17  mm, while the mean Euclidean distance for MobileEYE is 17.76 $\pm$ 7.69 mm. This is a 7.44\% increase in the error for the MobileEYE compared to the Tobii Pro Nano eye tracker. The lower minimum error for the Tobii Pro Nano suggests it can achieve higher accuracy in certain conditions compared to the MobileEYE.

\begin{table}[ht]
\centering
\caption{Overall errors for MobileEYE and Tobii Pro Nano (n = 582 recordings).}
\begin{tabular}{lll}
\hline
Statistic &  Tobii Pro Nano 
&MobileEYE \\
\hline
Mean &  16.53 &17.76 \\
Standard Deviation &  7.17 &7.69 \\
Min &  4.16 &9.05 \\
Max &  80.56 &53.45 \\
\hline
\end{tabular}
\label{tab:overall_error}
\end{table}



MobileEYE demonstrated a slightly higher median error than the Tobii Pro Nano. The lower maximum error in MobileEYE may be influenced by its model design, which constrains predictions within predefined screen boundaries. To evaluate whether the observed differences in error distributions were statistically meaningful, the Wilcoxon signed-rank test was conducted. The test result ($W = 65518.0, p = 0.278$) indicated no statistically significant difference in the median error values between the two methods.

\subsection{Null Values and Detection Failures}

One of the limitations observed in the results is the presence of null values in both eye-tracking techniques. Null values occur when the eye-tracking system fails to detect and record the gaze data for certain frames. Understanding the reasons behind these detection failures is essential for improving the reliability of both eye-tracking technologies.

According to the results, the Tobii Pro Nano eye tracker had a null value percentage of 28.38\%, while the MobileEYE had a lower null value percentage of 12.04\%. The high number of gaze detection failures in the Tobii Pro Nano eye tracker is mainly due to calibration issues and individual participant differences. Inaccurate calibration can lead to detection failures. For example, if the calibration process does not fully capture the participant's eye characteristics, the Tobii eye tracker tries to predict the gaze locations using the default calibration. However, it returns null values instead of providing inaccurate gaze data when it fails to predict reliable gaze coordinates. This approach helps avoid the risk of false predictions, but it also results in a higher percentage of null values. The main reason for not detecting any gaze data with the MobileEYE is that it failed to detect any faces in the frame. This is mainly because of the lighting conditions and blurriness of the recorded videos. 

Another factor that can affect gaze detection is the participant's characteristics, such as height and the use of prescription glasses. Prescription glasses that block infrared light can interrupt the eye tracker's ability to detect and track the eyes. Additionally, a participant's height may affect the angle at which the eye tracker and smartphone are positioned. If the eye tracker cannot properly see the participant's eyes, calibration may fail, resulting in a high number of null values.

\subsection{Calibration Failures}
\label{cali_failed}

The five-point calibration method was used to calibrate the Tobii Pro Nano. Calibration points were displayed as red dots in the four corners of the screen and the centre of the screen. Participants fixated on one blinking dot at a time. The dwell time per calibration point was 3 seconds, and a maximum of 3 attempts per calibration point. If any one of the calibration points failed to calibrate within the three attempts allowed at each point, participants had to restart the calibration from the beginning. Each participant was allowed five attempts per recording to calibrate all the points successfully. During these attempts, the table's height and the eye tracker's angle were changed to ensure the eye tracker could clearly see the participant's eye. If the eye tracker could not be calibrated within these attempts, the default calibration settings were used.

\begin{figure}[ht]
  \centering 
  \includegraphics[width=0.70\textwidth]{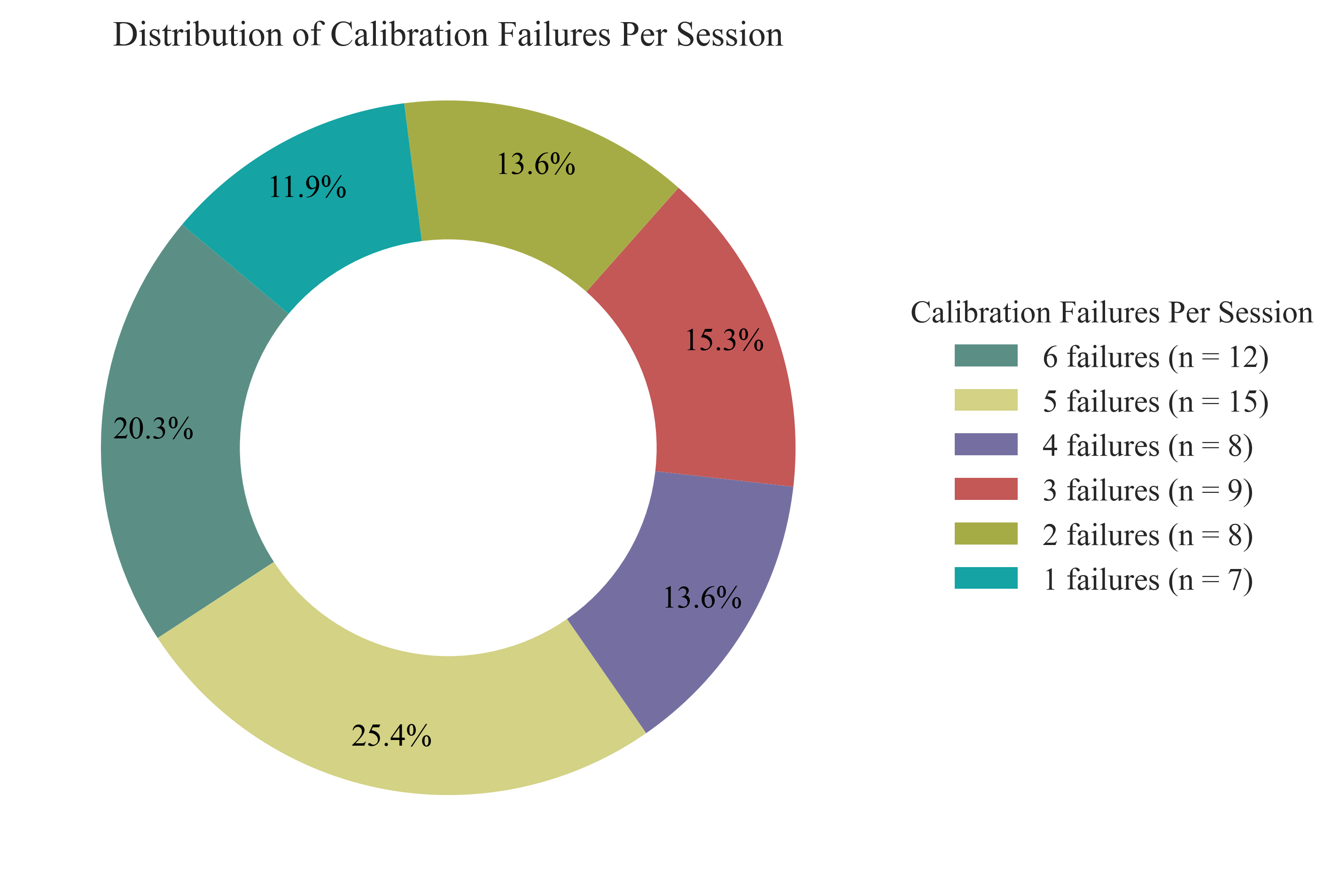}
  \caption{Distribution of calibration failures per session. Each session consists of 6 recordings. The segments indicate the number of recordings with calibration failures and had to proceed with the default calibration}
  \label{fig:calibration_failure}
\end{figure}

According to the results, 45.5\% of the sessions had no calibration failures, indicating successful calibration on the first attempt. Most of the sessions (54.5\%) had at least one calibration failure and after multiple attempts, these calibrations were successful. However, 10.05\% of the recordings had to proceed with the default calibration of the Tobii Pro Nano as they failed to calibrate successfully. Figure \ref{fig:calibration_failure} illustrates the distribution of the calibration failures per session.

Calibration was unsuccessful in every attempt across all 12 recordings for two sessions, both belonging to the same participant. The likely cause of this failure was the use of cosmetic eyelash extensions, which may have obstructed the pupil and interfered with the Tobii Pro Nano eye tracker's performance. In the remaining 47 recordings where calibration also failed, the specific causes could not be determined. No consistent correlation was observed with factors such as age, prescription eyeglasses, or mobile device position.

Calibration is a time-consuming process, which complicates the eye-tracking process. When the calibration is successful in the first attempt for all the recordings, the average duration for the entire process for a participant is 32 minutes and 37 seconds. However, this duration increases to 38 minutes and 04 seconds when there is a single calibration failure. This time includes the waiting period for the application to calibrate and additional time is required to adjust the height of the table and the angles of the eye tracker to ensure a clear view of the participant's pupil.

Due to these reasons, researchers are moving into developing calibration-free eye trackers \cite{garcia2023calibration, mokatren2024calibration}. MobileEYE does not require calibration and could be a potential avenue for calibration-free eye tracking. MobileEYE can estimate the gaze location without manual calibration, as the deep learning model can adapt to variations in eye shapes, lighting and head positions.

\subsection{Age}

In this experiment, data were collected from participants aged 18 to 68 years old. Table \ref{tab:age_group_statistical_measures} presents the statistical measures of errors for the MobileEYE and Tobii Pro Nano across different age groups. According to the results, the lowest mean error for MobileEYE was reported in the 39-38 age group, which is 14.96 $\pm$ 6.68 mm. Mean error reported in age groups 18-28 and 29-38 were 15.04 $\pm$ 5.41 mm and 15.37 $\pm$ 4.37 mm. This indicates that the MobileEYE maintains consistent accuracy for younger participants.

The mean error of the MobileEYE has a significant increase for older age groups. The mean error in the 49-58 age group increased to 19.09 $\pm$ 17.52 mm, which represents a 27\% increase compared to the 29-38 age group. Additionally, the mean error in the 59-68 age group further increased to 31.06 $\pm$ 13.42 mm, which is approximately 102\% higher than the 29-38 age group. This is because only 9\% of the participants in the training dataset are older than 60 and 21\% are older than 50, leading to less accurate training and performance for these age groups.

\begin{table}[ht]
\centering
\caption{Age Group-wise Statistical Measures of Errors for MobileEYE and Tobii Pro Nano. (Error in mm). n = number of recordings per age group.}
\begin{tabular}{llcccccc}
\hline
Statistic& Error Type & 18-28 & 29-38 & 39-48 & 49-58 & 59-68 \\
& & (n=102) & (n=258) & (n=108) & (n=72) & (n=42) \\
\hline
Mean & MobileEYE & 15.04 & 15.37 & 14.96 & 19.09 & 31.06 \\
 & Tobii Pro Nano & 17.89 & 15.25 & 17.80 & 17.52 & 16.37 \\
SD. & MobileEYE & 5.41 & 4.37 & 3.57 & 10.18 & 13.42 \\
 & Tobii Pro Nano & 9.54 & 6.33 & 6.68 & 6.58 & 5.33 \\
Min & MobileEYE & 9.05 & 10.32 & 9.82 & 9.99 & 11.70 \\
 & Tobii Pro Nano & 5.17 & 7.75 & 9.49 & 4.16 & 8.74 \\
Max & MobileEYE & 44.69 & 47.12 & 30.47 & 51.52 & 52.24 \\
 & Tobii Pro Nano & 80.56 & 68.91 & 64.75 & 45.65 & 29.07 \\
\hline
\end{tabular}
\label{tab:age_group_statistical_measures}
\end{table}

In contrast, the Tobii Pro Nano exhibits a different trend, likely due to its more extensive training data across various age groups, leading to a more consistent performance regardless of the participant's age. The mean error for the Tobii Pro Nano is highest in the 18-28 age group at 17.89 $\pm$ 9.54, which then decreases by 14.7\% in the 29-38 age group to 15.25 $\pm$ 6.33. The error remains relatively consistent in the 39-48 and 49-58 age groups with only slight variations, being 0.5\% lower and 2\% lower, respectively, compared to the 18-28 age group. Finally, in the 59-68 age group, the mean error decreases by 8.5\% to 16.37 $\pm$ 5.33, indicating a more stable performance across different age groups compared to the MobileEYE.

\begin{figure}[ht]
  \centering 
  \includegraphics[width=0.70\textwidth]{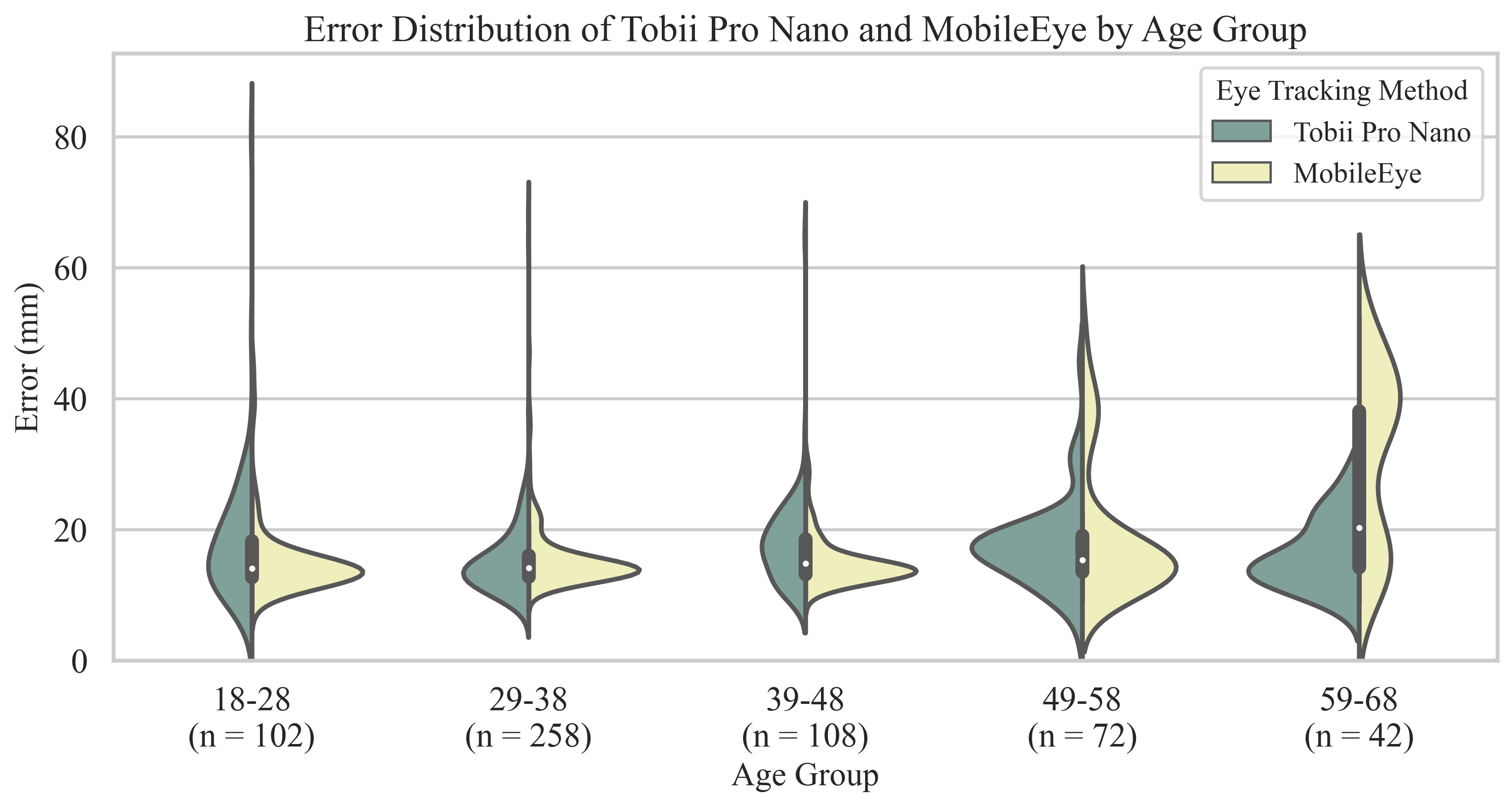}
  \caption{Error distribution of Tobii Pro Nano and MobileEYE by age group. n = number of recordings per age group.}
  \label{fig:error_vs_age}
\end{figure}

Figure \ref{fig:error_vs_age} illustrates a visual representation of the distribution of errors for the Tobii Pro Nano and MobileEYE across different age groups. For younger age groups, the Tobii Pro Nano shows a long tail extending towards higher errors, indicating the presence of outliers. In contrast, the MobileEYE shows a more concentrated distribution around the median, with fewer high-error outliers. As age increases, the MobileEYE exhibits increased variability, particularly in the 49-58 and 59-68 age groups. The Tobii Pro Nano maintains a more consistent error distribution across all age groups. However, the variability of the Tobii Pro Nano eye tracker increased with older age participants. According to the ANOVA test results, the effect of age is highly significant for the error of the eye tracking method, with an F-value of 16.82 and a p-value $<$0.001.

One limitation of this evaluation is the inability to provide sufficient data to assess each age group equally. For example, there were only 10 participants in the older groups (ages 49 to 68) and 30 participants in the age group from 18 to 38. However, these results highlight the importance of considering age-related factors in the design of eye-tracking devices to ensure accuracy and consistency across different age groups.

\subsection{Gender}

Table \ref{tab:gender_vs_error} compiles the mean error and standard deviation for the Tobii Pro Nano and Mobile Eye Tracking methods across two different genders. The mean error for the females was higher in MobileEYE as 70\% of the older participants (age 49-68) were females. Tobii Pro Nano has a similar mean error for both males and females, with 16.63 mm $\pm$ 6.52 mm for males and 16.42 mm $\pm$ 7.78 mm for females.

\begin{table}[ht]
\centering
\caption{Gender-wise Statistical Measures of Errors for MobileEYE and Tobii Pro Nano. (Error in mm).  n = number of recordings per gender.}
\begin{tabular}{lllcc}
\hline
 &  \multicolumn{2}{c}{Male (n=288)
}&\multicolumn{2}{c}{Female (n=294)}\\
 &  MobileEYE & Tobii Pro Nano
&MobileEYE & Tobii Pro Nano \\ 
\hline
Mean &  15.79 & 16.42 
&17.73 & 16.63 \\ 
SD &  5.18 & 7.78 
&9.26 & 6.52 \\ 
Min &  9.82 & 4.16 
&9.05 & 8.30 \\ 
Max &  47.12 & 80.56 &52.24 & 64.75 \\ 
\hline
\end{tabular}
\label{tab:gender_vs_error}
\end{table}

\begin{figure}[ht]
  \centering 
  \includegraphics[width=0.70\textwidth]{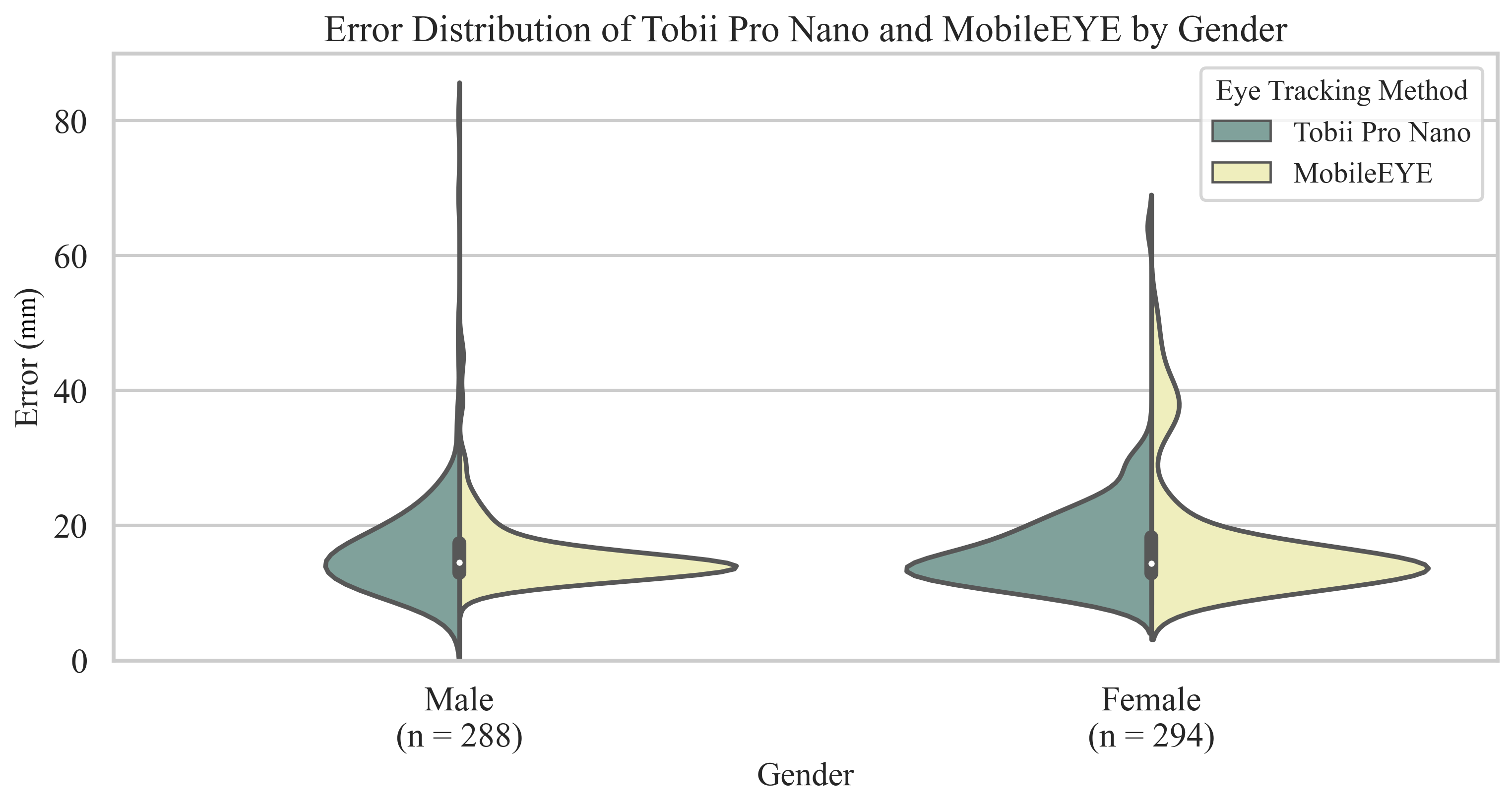}
  \caption{Error distribution of Tobii Pro Nano and MobileEYE by gender. n = number of recordings per gender.}
  \label{fig:error_vs_gender}
\end{figure}

The violin plot in Figure \ref{fig:error_vs_gender} visualises the error distribution for both male and female participants. This graph shows that both MobileEYE and Tobii Pro Nano have a similar central tendency in their error distributions for both genders. The ANOVA results also prove that the effect of gender is not significant, with an F-value of 6.377 and a p-value of 0.012.

\subsection{Vision Correction}

The impact of vision correction on the accuracy of eye-tracking was evaluated by considering participants who used either contact lenses or prescription glasses, as well as those who did not use any vision correction. Table \ref{tab:vision_correction} summarises the mean error and standard deviation for the Tobii Pro Nano and Mobile Eye Tracking methods across these categories. The results indicate that the type of vision correction used by participants affects the accuracy of both the Tobii Pro Nano and MobileEYE tracking methods. For participants wearing contact lenses, the mean error for Tobii Pro Nano was 12.46 $\pm$ 1.71 mm, while MobileEYE had an error 11.3\% higher at 13.87 $\pm$ 2.41 mm. For participants without any vision correction, the mean error for Tobii Pro Nano was 16.54 mm $\pm$ 7.60 mm and for MobileEYE, the mean error was 16.00 mm $\pm$ 6.85 mm.

\begin{table}[ht]
    \centering
    \caption{Comparison of Mean Error and Standard Deviation by Vision Correction Type (NVC - No Vision Correction, CL - Contact Lenses).}
    \begin{tabular}{lccccc}
        \hline
        \multirow{2}{*}{Type} & \multirow{2}{*}{n} & \multicolumn{2}{c}{Tobii Pro Nano} & \multicolumn{2}{c}{Mobile Eye Tracking} \\
        & & Mean Error & SD & Mean Error & SD \\ \hline
        CL & 24 & 12.46 & 1.71 & 13.87 & 2.409 \\ 
        Glasses & 198 & 17.07 & 6.62 & 18.54 & 8.85 \\ 
        NVC  & 360 & 16.54 & 7.60 & 16.00 & 6.85 \\ 
        \hline
    \end{tabular}
    \label{tab:vision_correction}
\end{table}

Participants using prescription glasses exhibited higher mean errors. The Tobii Pro Nano had a mean error of 17.07 $\pm$ 6.62 mm  which is 3.2\% higher than its error for participants without vision correction. MobileEYE showed a mean error of 18.54 mm $\pm$ 8.85 mm which is 15.9\% higher than its error for participants without vision correction. This increase in error can be attributed to the reflections and distortions caused by the glasses, which can obscure the eye-tracker's view. Also, calibration process failed continuously when the participant was wearing prescription glasses, which could block the infrared lights.

\begin{figure}[ht]
  \centering 
  \includegraphics[width=0.70\textwidth]{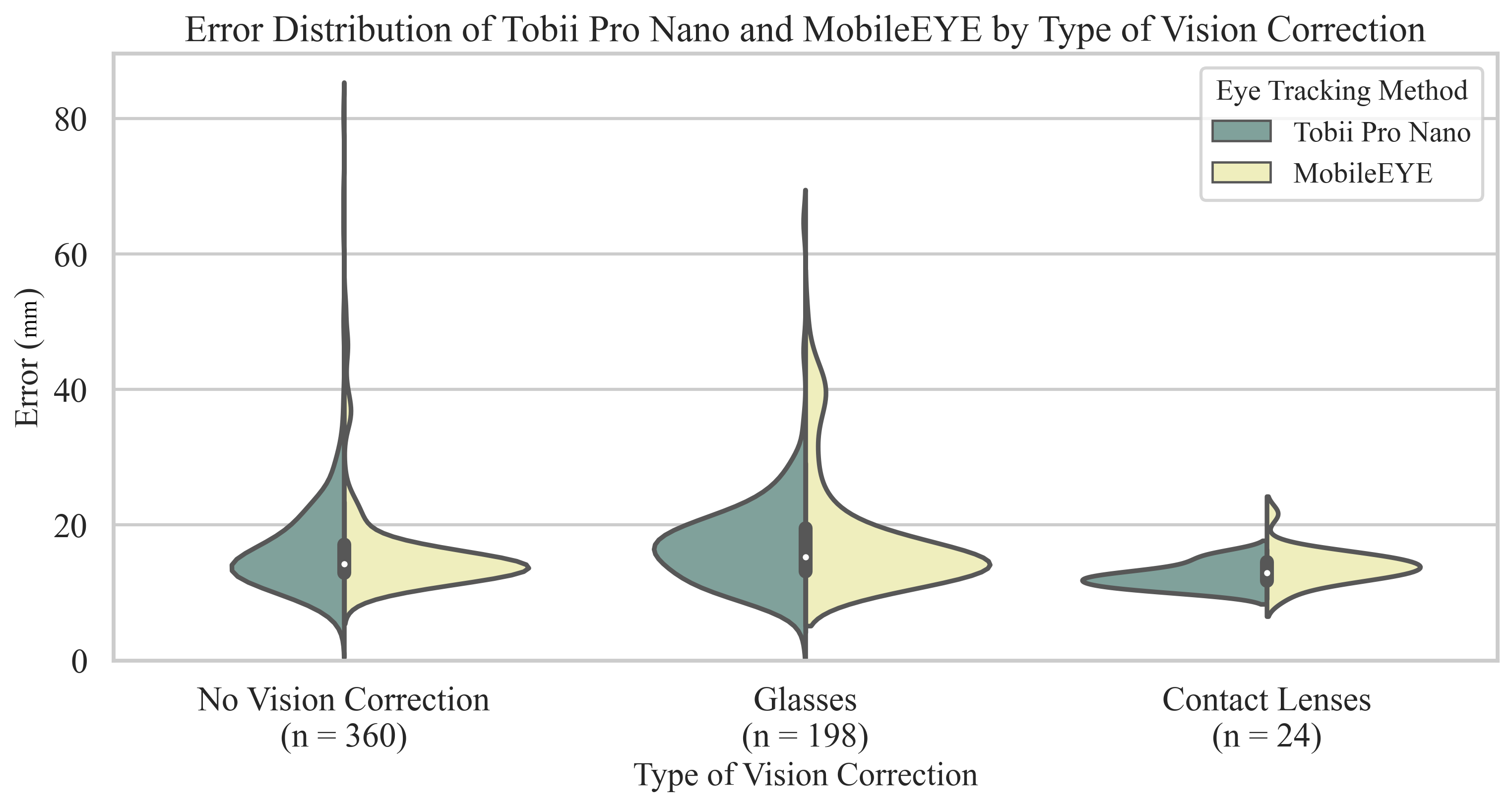}
  \caption{Error distribution of Tobii Pro Nano and MobileEYE by type of vision correction. n = number of recordings per vision correction type.}
  \label{fig:error_vs_glasses}
\end{figure}

Figure \ref{fig:error_vs_glasses} shows the error distribution for Tobii Pro Nano and MobileEYE across different types of vision correction. The violin plot shows that both eye-tracking methods exhibit higher error variability for participants wearing glasses compared to those using contact lenses or no vision correction. This visual representation confirms the statistical findings.

ANOVA results suggest that the effect of vision correction is highly significant, with an F-value of 11.377 and a p-value $<$ 0.001. The interaction between error type and vision correction is marginally significant, with an F-value of 2.447 and a p-value of 0.087, indicating that the difference in errors between MobileEYE and Tobii Pro Nano may vary depending on the type of vision correction used by participants.

These results underscore the importance of considering vision correction types when evaluating the accuracy of eye-tracking systems. The presence of contact lenses and without any vision correction results in relatively lower errors compared to glasses. This understanding can guide improvements in future appearance-based eye-tracking algorithms to better accommodate users with different vision correction needs.

\subsection{Eye Position}

Evaluation of the performance across these three eye positions provides insights into how different viewing angles and head positions impact the accuracy of the eye-tracking methods. The statistical measures of errors for the MobileEYE and Tobii Pro Nano tracking methods across these positions are summarised in Table~\ref{tab:device_holding_positions}.

\begin{table}[ht]
\centering
\caption{Statistical Measures of Errors for MobileEYE and Tobii Pro Nano by Device Holding Positions (Error in mm)}
\begin{tabular}{llccccccc}
\hline
Method & Direction & n & Mean & SD & Min & Max \\
\hline
\multirow{3}{*}{Tobii Pro Nano} & Looking Down & 194 & 17.81 & 8.62 & 7.24 & 80.56 \\
 & Looking Straight& 194 & 16.00 & 7.23 & 5.17 & 64.75 \\
 & Looking Up& 194 & 15.69 & 4.89 & 4.16 & 34.54 \\
\hline
\multirow{3}{*}{MobileEYE} & Looking Down& 194 & 17.26 & 7.01 & 11.04 & 49.33 \\
 & Looking Straight& 194 & 17.69 & 7.32 & 9.05 & 52.24 \\
 & Looking Up& 194 & 18.33 & 8.65 & 10.35 & 53.45 \\
\hline
\end{tabular}
\label{tab:device_holding_positions}
\end{table}

According to these results, when the participants were looking down, the mean error for MobileEYE was 17.26 $\pm$ 7.01 mm. In contrast, for Tobii Pro Nano, the mean error was slightly higher at 17.81 $\pm$ 8.62 mm. The error for Tobii Pro Nano in this position was 7.7\% higher than its overall average error of 16.53 $\pm$ 7.17. This indicates that when the eye levels are above the Tobii Pro Nano, higher errors tend to be observed due to the difficulty in tracking when the participant's gaze is directed downward.

For the looking straight position, the mean error for MobileEYE was 17.69 $\pm$ 7.32 mm which was 10.6\% higher compared to Tobii Pro Nano's error of 16.00 $\pm$ 7.23 mm. When looking up, the mean error for MobileEYE increased to 18.33 $\pm$ 8.65 mm. In contrast, Tobii Pro Nano had a lower mean error of 15.69 $\pm$ 4.89 mm. This represents a 17.7\% higher error for MobileEYE compared to Tobii Pro Nano in the looking-up position. The increased error for MobileEYE when looking up suggests that the upper eyelids and eyelashes obstruct the view of the eyes, leading to less accurate tracking.



The results from the ANOVA analysis further support these observations. The analysis shows that the interaction between the error type and eye level is statistically significant, with an F-value of 4.53 and a p-value of 0.011, indicating that the effect of eye level on the error is different for MobileEYE and Tobii Pro Nano. Tobii Pro Nano had the lowest mean error when looking up (15.69 mm), while MobileEYE had the lowest mean error when looking down (17.26 mm). This indicates that each method has its optimal performance position.

\subsection{Device Type}

Different smartphones are equipped with different hardware configurations, including camera resolutions, screen sizes, processing capabilities and sensor qualities. These variations can impact the performance of eye-tracking algorithms. The statistical measures of errors for the MobileEYE and Tobii Pro Nano tracking methods across these two devices are summarised in Table \ref{tab:device_vs_error}. 

\begin{table}[ht]
\centering
\caption{Statistical Measures of Errors for MobileEYE and Tobii Pro Nano by Device Type (Error in mm)}
\begin{tabular}{llccccc}
\hline
Device & Method & n &Mean & SD & Min & Max \\
\hline
\multirow{2}{*}{Samsung S22} & MobileEYE & 282 &18.66 & 8.05 & 10.35 & 53.45 \\
& Tobii Pro Nano & 282 & 16.80 & 6.32 & 4.16 & 63.66 \\
\hline
\multirow{2}{*}{iPhone 14} & MobileEYE & 300 & 16.92 & 7.26 & 9.05 & 47.30 \\
& Tobii Pro Nano & 300 & 16.28 & 7.86 & 5.17 & 80.56 \\
\hline
\end{tabular}
\label{tab:device_vs_error}
\end{table}

The results show that MobileEYE had a 10.3\% higher mean error on Samsung devices (18.66 $\pm$ 8.05 mm) compared to iPhone devices (16.92 $\pm$ 7.26 mm). While higher-resolution cameras can potentially provide more detailed images, improving the accuracy of appearance-based eye-tracking algorithms, this relationship is complex and not solely dependent on resolution. The actual video resolution used for analysis is often lower than the camera's raw resolution and other factors, such as image processing algorithms and sensor quality, play important roles in the accuracy of the gaze estimation \cite{cheng2024appearance}.


Figure \ref{fig:collected_data} illustrates the difference between the videos recorded from the iPhone 14 and Samsung S22. The videos recorded using the Samsung S22 contain blurry frames, which could be attributed to several factors, including the limitations of the auto-focus (AF) algorithms, which may struggle with high contrast patterns in participants' clothing, reflective materials, or challenging lighting conditions. Additionally, differences in camera firmware decisions related to ISO (sensitivity) and shutter speed trade-offs may have contributed to motion blur and reduced image sharpness despite both phones being tested under the same conditions.

As expected, the mean error for Tobii Pro Nano remains consistent across these two decide types. The mean error for Tobii Pro Nano on Samsung devices is only 3.1\% higher than on iPhone devices. This minimal difference is due to the calibration process of the eye tracker, which relies on the size of the smartphone screen rather than the camera resolution. Therefore, smartphone device characteristics have minimal impact on commercial eye tracker's accuracy.

\begin{figure}[ht]
  \centering 
  \includegraphics[width=0.60\textwidth]{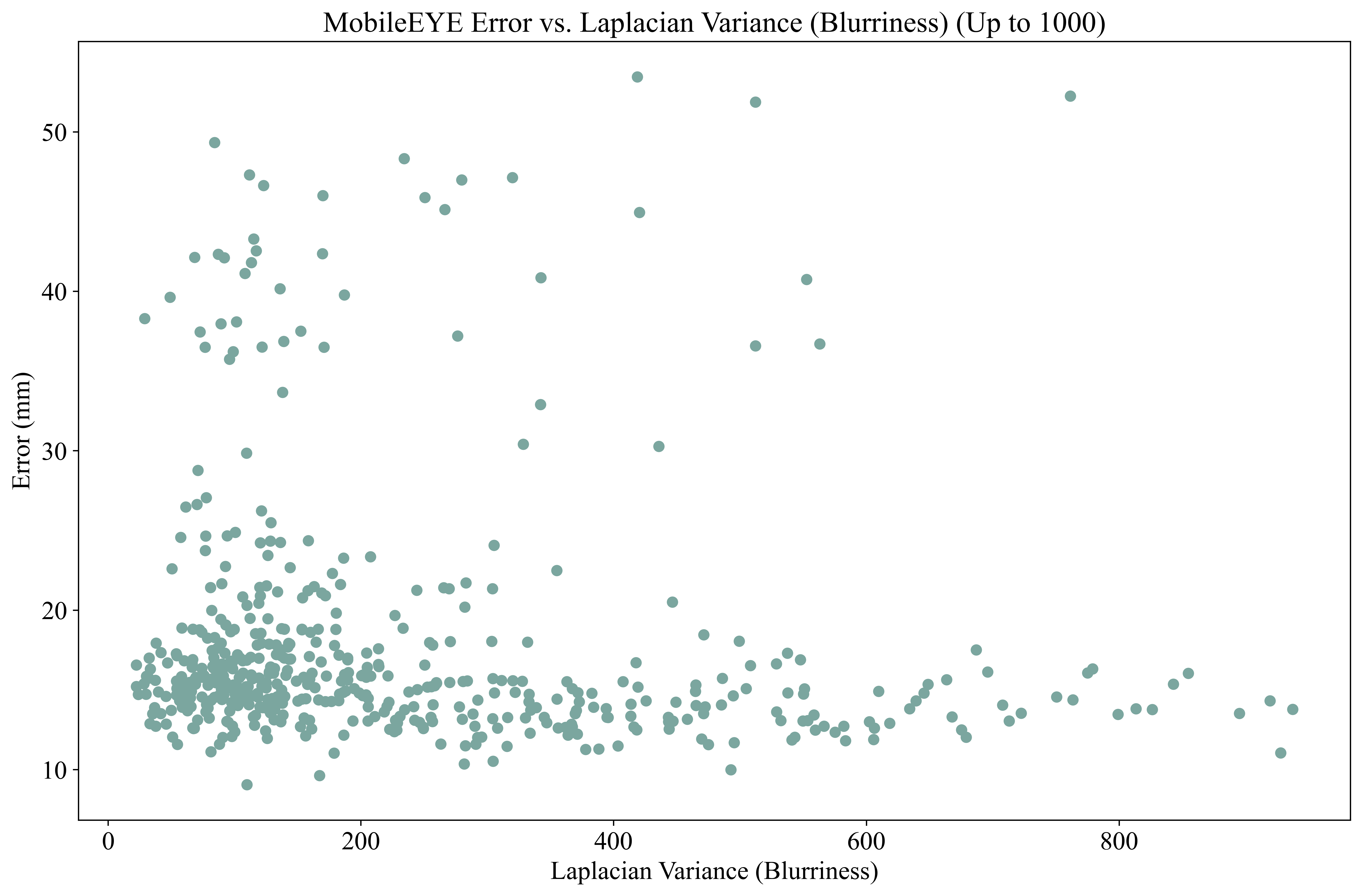}
  \caption{Error distribution for MobileEYE vs. Laplacian Variance (Blurriness) for values up to 1000. 21 outliers were removed to avoid skewing the analysis.}
  \label{fig:MobileEYE_error_blurriness}
\end{figure}

The error of the MobileEYE was further evaluated against the blurriness of the videos. Using the Laplacian variance method \cite{bansal2016blur}, the blurriness of each frame in the videos recorded on both Samsung and iPhone devices was measured. Laplacian variance is a metric used to determine the blurriness of a frame. It is calculated by applying the Laplacian operator to a frame to measure the variance of the Laplacian. High variance indicates a sharp image with fine details, while low variance indicates a blurred image. Figure \ref{fig:MobileEYE_error_blurriness} shows the relationship between the error in MobileEYE and the mean Laplacian variance (up to 1000) of the video.

The plot reveals a broad distribution of error values across different levels of blurriness. While there is a concentration of data points with relatively low error (ranging from 10 to 20 mm) at lower Laplacian variance values (indicating more blurred frames), the data also show that higher variance (indicating sharper images) does not consistently correlate with lower errors. Some high-variance points still exhibit relatively high errors. This suggests that while image sharpness may have some influence on the error rate, other factors likely contribute to the overall accuracy of the MobileEYE algorithm. The lack of a clear trend indicates the complexity of the relationship between image blurriness and the accuracy needing further investigation into additional variables that may impact the accuracy of gaze estimation.

However, ANOVA results suggest that the type of device used significantly influences the error measurements with an F-value of 6.70 and a p-value $<$ 0.001. These results highlight the importance of considering device-specific factors when developing an appearance-based eye-tracking algorithm. While creating device-specific models would undermine the goal of generalisability, a potential strategy could involve incorporating device-specific calibration steps or applying transfer learning techniques that adapt the model to different devices without requiring entirely separate models. Another approach could be to include a diverse set of device characteristics during training, ensuring the model learns to generalise across various devices while maintaining accuracy.

\subsection{Lighting Conditions}

\begin{figure}[ht]
    \centering
    \begin{subfigure}[b]{0.48\textwidth}
        \centering
        \includegraphics[width=\textwidth]{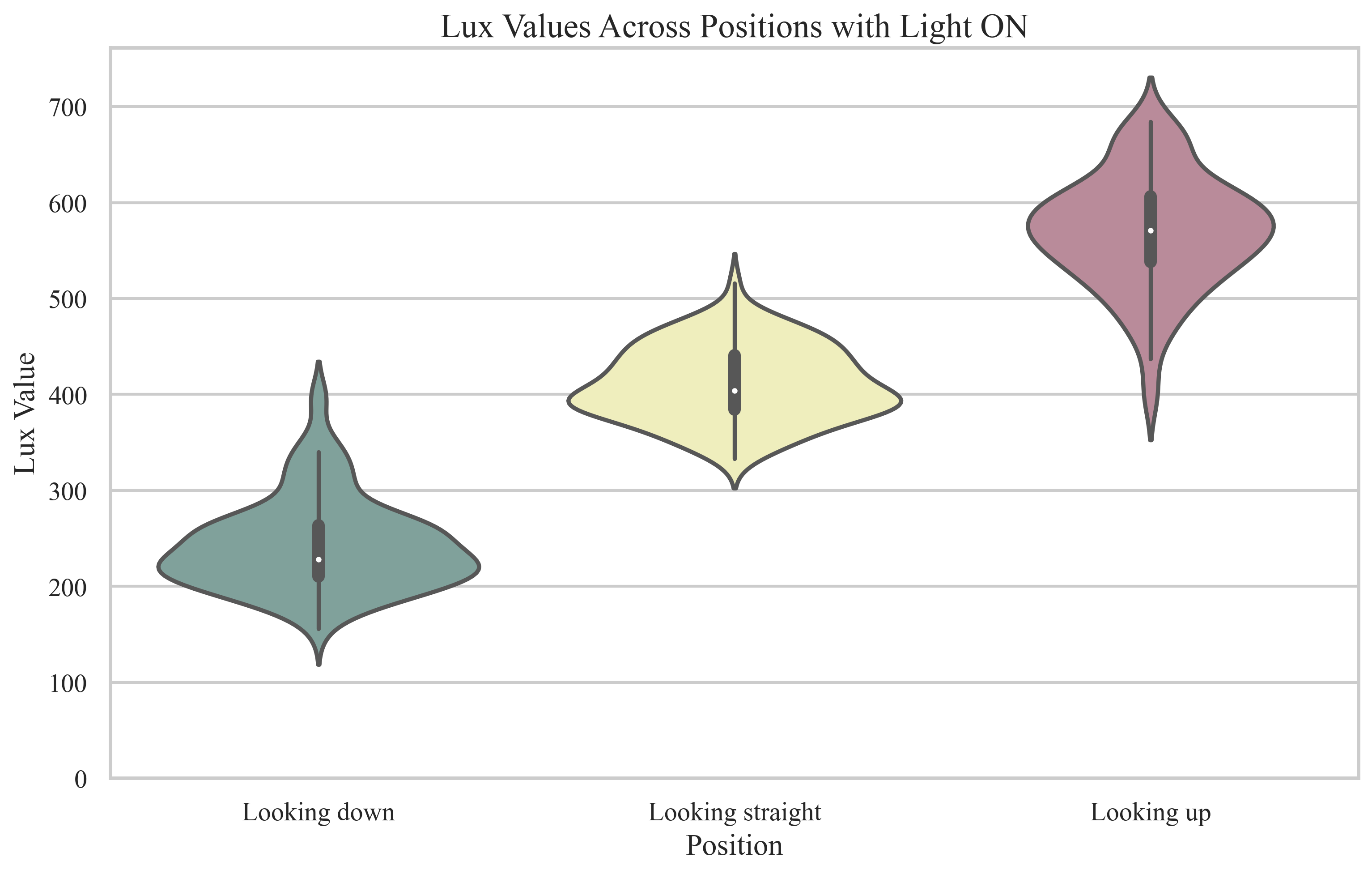}
        \caption{Lights ON}
        \label{fig:lights_on}
    \end{subfigure}
    \hfill
    \begin{subfigure}[b]{0.48\textwidth}
        \centering
        \includegraphics[width=\textwidth]{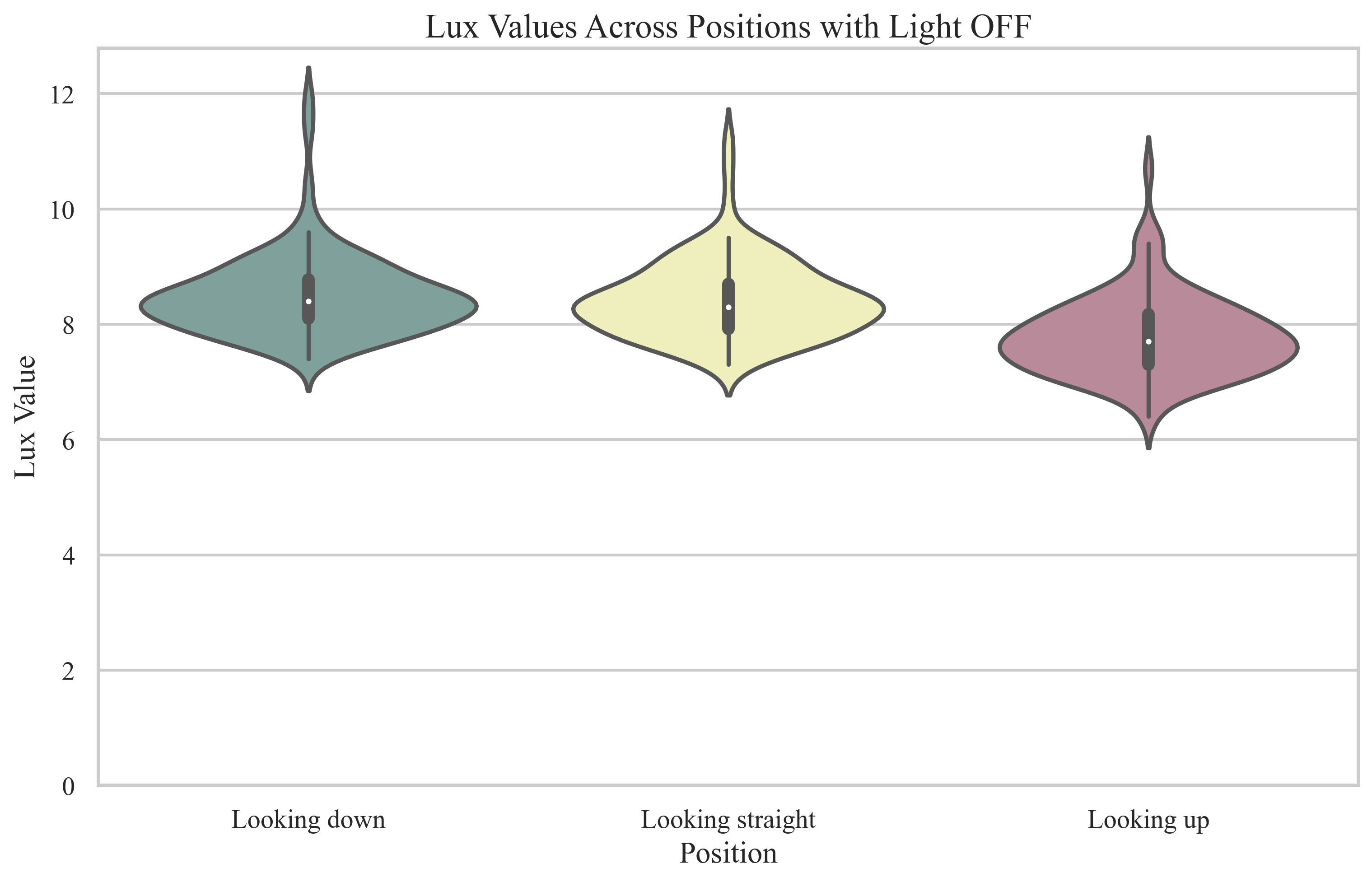}
        \caption{Lights OFF}
        \label{fig:light_off}
    \end{subfigure}
    \hfill
    \caption{Lux Values Across Positions with Lights ON  and OFF.}
    \label{fig:lux_values}
\end{figure}

During the data collection, the amount of light reaching participants' faces was measured using a lux meter. Therefore, the lux values rely on the head position. Figure \ref{fig:lights_on} illustrates the recorded lux values across three positions when the lights are ON, while Figure \ref{fig:lights_on} shows the lux values when the lights are OFF. 

Lighting conditions were categorised as \say{Low Light} when the Lux $<$ 15 and \say{High Light} when the Lux $>$ 150. The collected lux values show distinct differences based on the light conditions and gaze angles. In low light conditions, the mean lux values are 8.49 $\pm$ 0.70 when looking down, 8.42 $\pm$ 0.67 when looking straight and 7.78 $\pm$ 0.69 when looking up. In high light conditions, the mean lux values are 243.14 $\pm$ 47.08 when looking down, 410.70 $\pm$ 38.56 when looking straight and 569.06 $\pm$ 58.52 when looking up.

The lux values are highest when participants are "Looking up" because, in that position, more light from the ceiling directly reaches their faces. However, when participants are "Looking down," their faces are less exposed to direct ceiling light, likely due to shadowing effects created by their own facial features and the angle of their heads, resulting in lower lux values. When the lights are OFF, the lux values are lower and more uniform across different gaze angles, likely due to the presence of diffuse ambient light, which is less directional and, therefore, remains consistent regardless of the participant's gaze direction.

\begin{table}[ht]
\centering
\caption{Error Analysis by Light Condition for MobileEYE and Tobii Pro Nano. (Error in mm)}
\begin{tabular}{lccccc}
\hline
Light Condition & Method & Mean & SD & Min & Max \\
\hline
\multirow{2}{*}{Low Light} & MobileEYE & 18.01 & 8.05 & 12.82 & 51.52 \\
          & Tobii Pro Nano & 16.17 & 6.68 & 4.16 & 68.91 \\
\hline
\multirow{2}{*}{High Light} & MobileEYE & 15.81 & 6.03 & 9.05 & 52.24 \\
           & Tobii Pro Nano & 16.88 & 7.62 & 7.10 & 80.56 \\
\hline
\end{tabular}
\label{tab:light_condition_error}
\end{table}

Table \ref{tab:light_condition_error} summarises the statistical measures of errors for the MobileEYE and Tobii Pro Nano across two lighting conditions. The results indicate that MobileEYE had a 13.9\% higher mean error in low light conditions (18.01 $\pm$ 8.05 mm) compared to high light conditions (15.81 $\pm$ 6.03 mm). This suggests that MobileEYE's appearance-based eye-tracking algorithm is more sensitive to lighting conditions because it depends on visible light to capture facial features. In contrast, Tobii Pro Nano showed a relatively consistent mean error across both lighting conditions, with only a 4.4\% increase in mean error from low light (16.17 $\pm$ 6.68 mm) to high light (16.88 $\pm$ 7.62 mm).

\begin{figure}[ht]
  \centering 
  \includegraphics[width=0.70\textwidth]{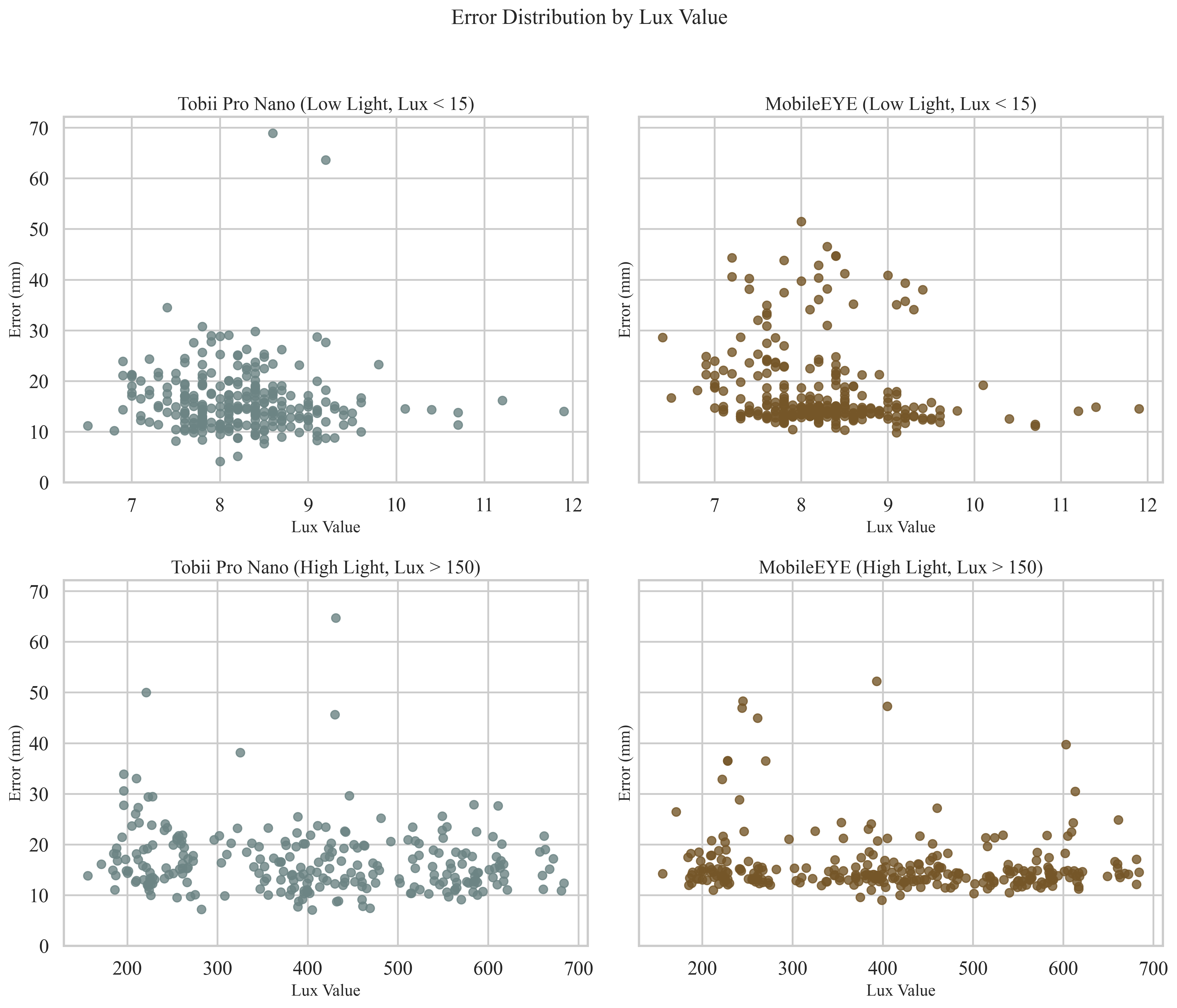}
  \caption{Error distribution by lux value for Tobii Pro Nano and MobileEYE with high light (lux $>$ 150) and low light (lux $<$ 15)}
  \label{fig:error_vs_lux}
\end{figure}

The ANOVA test results for Tobii Pro Nano show an F-statistic of 1.29 and a p-value of 0.26, indicating no significant difference in error rates between low and high light conditions. In contrast, the ANOVA test results for MobileEYE indicate an F-statistic of 13.88 and a p-value of 0.0002, demonstrating a significant difference in error rates between low and high light conditions. These findings are further illustrated in Figure \ref{fig:error_vs_lux}, which illustrates the error distribution for both the Tobii Pro Nano and MobileEYE eye-tracking methods under two lighting conditions (low light and high light).

For Tobii Pro Nano, under low lighting conditions, the errors are widely spread, with few outliers reaching up to 70mm. For the MobileEYE, the error distribution under low light conditions also shows a broad spread. Especially when the lux value is less than 7, errors exceed 20 mm. As the lux value increases, MobileEYE demonstrates a significant decrease in error, highlighting its sensitivity to improved lighting conditions.

In the high light, the Tobii Pro Nano's errors are more uniformly distributed across the range of lux values, with the majority of errors between 10 and 30 mm. Tobii Pro Nano relies on infrared illumination instead of visible light to track eye movements. This allows the eye tracker to maintain accuracy and stability even in different visible light environments.

MobileEYE also shows increased stability with more outliers in the high light. The presence of some high-error instances suggests that other factors may still impact accuracy. Therefore, this analysis suggests that ensuring proper lighting conditions is important for achieving optimal accuracy in appearance-based eye-tracking methods.

\section{Conclusion}
\label{conclusion}

This study evaluated a deep learning-based gaze estimation algorithm, MobileEYE, by comparing its performance with the Tobii Pro Nano eye tracker across 582 recordings from 51 participants. The evaluation considered several sensitivity parameters, including age, gender, lighting conditions, device type, head position, and vision correction. MobileEYE produced a mean Euclidean error of 1.776 cm, compared to 1.653 cm for the Tobii Pro Nano, corresponding to an average increase of 0.123 cm. MobileEYE recorded fewer detection failures overall, as it does not rely on infrared-based calibration. However, higher error values were observed under low-light conditions and with image blurriness. Age group analysis showed an increase in error with older participants. Device-specific differences were also noted, with varying performance between the Samsung S22 and iPhone 14. Higher errors were observed in recordings from participants wearing prescription glasses. Calibration failures occurred in 61 out of 582 recordings. Of these, 12 recordings were from a single participant where calibration failed in all attempts, possibly due to cosmetic eyelash extensions. The remaining 49 calibration failures were distributed across other recordings, but no clear correlation was identified with specific participant characteristics or conditions.

Several limitations have been identified in this study. The representation of participants was uneven across age groups, particularly in the older categories, which may have influenced the observed error trends. Calibration issues with the Tobii Pro Nano resulted in the use of default calibration settings in several sessions, potentially affecting the accuracy of the baseline measurements. Additionally, the study was limited to static positioning of devices during data collection, which may not reflect the full range of real-world mobile usage scenarios. Future work may focus on addressing these limitations by incorporating more diverse participant demographics and expanding device coverage to include a wider range of smartphone models and eye-tracking systems. Improving the balance across age groups and vision correction types would allow more granular analysis. Further efforts can explore dynamic usage scenarios that reflect real-world conditions, such as hand-held interactions and outdoor environments. Optimising the MobileEYE algorithm for real-time performance, reducing its sensitivity to lighting and image quality, and investigating adaptive or device-specific calibration strategies may enhance its robustness and applicability across diverse settings.

\section*{CRediT authorship contribution statement}

\textbf{Nishan Gunawardena:} Conceptualisation, Methodology, Software, Investigation, Writing-Original draft. \textbf{Gough Yumu Lui:} Conceptualisation, Methodology, Writing-Reviewing and Editing, Supervision. \textbf{Bahman Javadi:} Resources, Writing-Reviewing and Editing, Supervision. \textbf{Jeewani Anupama Ginige:} Conceptualisation, Methodology, Writing-Reviewing and Editing, Supervision, Project administration, Resources.

\section*{Declaration of competing interest}

The authors declare that they have no known competing financial interests or personal relationships that could have appeared
to influence the work reported in this paper.

\section*{Declaration of generative AI and AI-assisted technologies in the writing process}

During the preparation of this work the author used ChatGPT 4o in order to improve the readability and language of the manuscript. After using this tool, the authors reviewed and edited the content as needed and take full responsibility for the content of the published article.

\section*{Acknowledgements}

This research was supported by Western Sydney University through institutional funding and resources.

\section*{Data availability}

Data will be made available on request.

\bibliographystyle{elsarticle-num}

\end{document}